\documentclass[10pt,twocolumn,letterpaper]{article}

\usepackage[pagenumbers]{iccv} 

\definecolor{iccvblue}{rgb}{0.21,0.49,0.74}
\usepackage[pagebackref,breaklinks,colorlinks,allcolors=iccvblue]{hyperref}
\usepackage{multirow}
\usepackage{booktabs}
\usepackage{multirow}
\usepackage{wrapfig}
\usepackage{xcolor}
\usepackage{microtype}
\usepackage[accsupp]{axessibility}
\newcommand{\chinthani}[1]{\textcolor{black}{#1}}

\def\paperID{10630} 
\def\confName{ICCV}
\def\confYear{2025}

\title{IMoRe: Implicit Program-Guided Reasoning for Human Motion Q\&A}
\author{Chen~Li$^{1}$\thanks{Equal contribution.} \and Chinthani~Sugandhika$^{1,2}$\footnotemark[1]\and Yeo Keat Ee$^{1}$\and~Eric~Peh$^{1}$\and~Hao~Zhang$^{1}$\and~Hong~Yang$^{1}$~\and~Deepu~Rajan$^{2}$\and Basura~Fernando$^{1,2}$\\
$^{1}$Institute of High-Performance Computing, Agency for Science, Technology and Research, Singapore\\
$^{1}$Centre for Frontier AI Research, Agency for Science, Technology and Research, Singapore\\
$^{2}$College of Computing and Data Science, Nanyang Technological University, Singapore
}

\begin{document}
\maketitle
\begin{abstract}
\sloppy
Existing human motion Q\&A methods rely on explicit program execution, where the requirement for manually defined functional modules may limit the scalability and adaptability. To overcome this, we propose an implicit program-guided motion reasoning (IMoRe) framework that unifies reasoning across multiple query types without manually designed modules. Unlike existing implicit reasoning approaches that infer reasoning operations from question words, our model directly conditions on structured program functions, ensuring a more precise execution of reasoning steps. Additionally, we introduce a program-guided reading mechanism, which dynamically selects multi-level motion representations from a pretrained motion Vision Transformer (ViT), capturing both high-level semantics and fine-grained motion cues. The reasoning module iteratively refines memory representations, leveraging structured program functions to extract relevant information for different query types. Our model achieves state-of-the-art performance on Babel-QA and generalizes to a newly constructed motion Q\&A dataset based on HuMMan, demonstrating its adaptability across different motion reasoning datasets. 
Code and dataset are available at: 
{\small\url{https://github.com/LUNAProject22/IMoRe}}.

\end{abstract}    
\section{Introduction}
\label{sec:intro}

Understanding human motion has been a crucial challenge in computer vision, with wide applications in human-computer interaction and embodied AI. 
Significant progress has been made in human motion understanding, especially in action recognition and motion forecasting, while fine-grained human motion understanding and reasoning over subtle motions have been overlooked. Identifying body parts or the direction of movement involved in an action remains challenging due to the lack of high-quality annotated datasets, occlusions, motion ambiguities, variability in human poses, and the difficulty of modeling spatiotemporal dependencies at a fine-grained level. 
Recently, the human motion Q\&A task has been introduced in \cite{nspose} to address complex and fine-grained human motion understanding. Given a motion sequence and a corresponding question in natural language, this task requires multistep reasoning to predict various attributes of the motion sequence.

\begin{figure}
    \centering
    \includegraphics[width=\columnwidth]{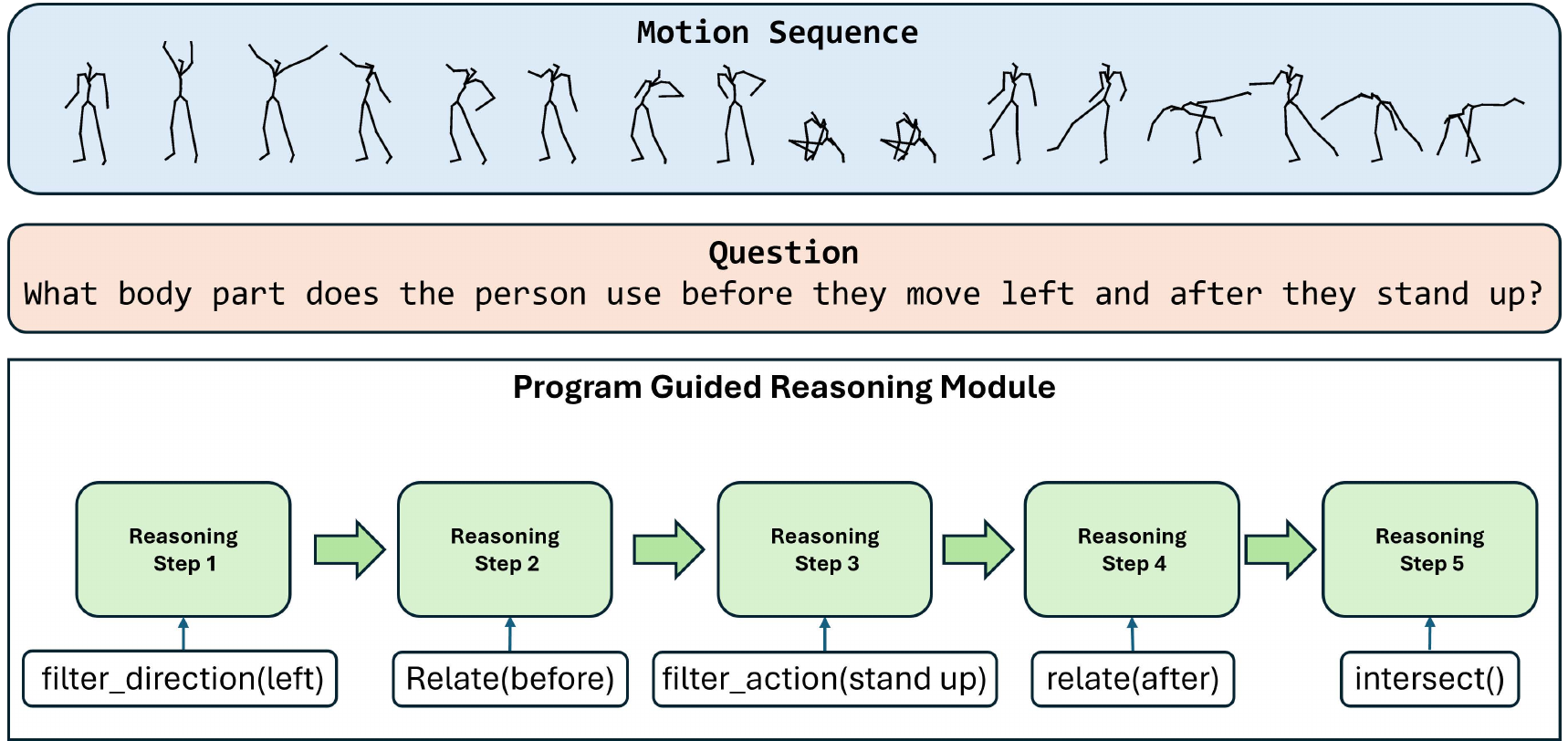}
        \caption{
    Illustration of the Program-Guided Reasoning Process for Human Motion Q\&A. Given a motion sequence and a natural language question, our model executes a structured reasoning process using a program-guided reasoning module.
    } 
    \label{fig:concept}
\end{figure}

Existing approaches \cite{nspose} to human motion question-answering (Q\&A) rely on predefined functional modules that explicitly perform reasoning steps such as filtering, relational inference, and querying. The advantage of explicit program-based reasoning compared to previous end-to-end approaches \cite{tevet2022motionclip} lies in its interpretability, as it provides access to all intermediate execution results, and in its data efficiency, since each functional module is designed with a small set of parameters.
Despite the advantages, these explicit program execution-based methods suffer from limitations in scalability and adaptability due to their reliance on manually defined functional modules. To overcome these constraints, we propose a novel implicit program-guided motion reasoning (IMoRe) framework that uses a unified reasoning module across multiple types of queries. Instead of leveraging separate handcrafted functional modules, our model dynamically adapts its reasoning process based on structured program functions, ensuring flexibility and generalization across diverse queries.

The specific design of our reasoning module is inspired by the Memory, Attention and Composition (MAC) \cite{mac}, which conducts iterative reasoning by composing control, read, and write units. The control unit identifies a series of operations from the given text, the read unit extracts relevant information from the visual input to execute each operation, and the write unit iteratively integrates the extracted information into the cell’s memory state, generating intermediate reasoning results. 
Although the MAC module infers reasoning operations through soft attention over question words, this approach can introduce ambiguity in operation generation, potentially affecting the accuracy of visual information retrieval. To solve this, our model is designed to condition directly on program functions, ensuring a more precise representation of reasoning operations. Furthermore, our approach enhances interpretability by explicitly defining operations, making the reasoning process more transparent.



To further enhance reasoning capacity, we introduce a program-guided reading mechanism. The motion understanding and reasoning task necessitates extracting different attributes from motion sequences. Furthermore, queries about action types require high-level motion sequence understanding, whereas queries about body parts require fine-grained information on localized motion. 
To address these challenges, we extract multi-level features from a pre-trained motion Vision Transformer (ViT) \cite{motionpatches} and allow the reasoning module to select the most appropriate feature level based on the program function. 
This design allows the model to access the most relevant information dynamically for each reasoning step. 

\begin{figure*}[th!]
    \centering
    \includegraphics[width=\textwidth]{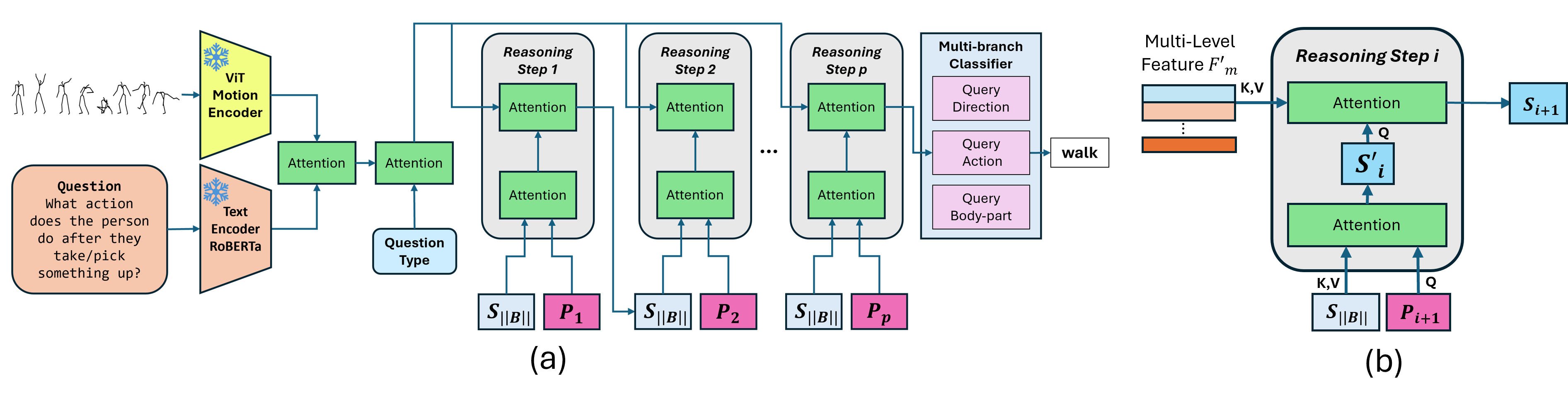}
        \caption{
(a) Overall pipeline of our Program-Guided Motion Reasoning Framework. Motion and text features are extracted using a ViT-based motion encoder and RoBERTa. These features are processed through a structured reasoning module, where intermediate memory states are iteratively refined using program guidance. A multi-branch classifier is adopted to predict the final answer. (b) Illustration of one reasoning step.
    } 
    \label{fig:Architecture}
\end{figure*}

We evaluate our approach on the Babel-QA benchmark, where our model significantly outperforms existing methods. 
To further test its generalization capability, we introduce a new motion Q\&A dataset based on the HuMMan dataset. Additionally, we assess the robustness of our model by using predicted programs, and simulating scenarios where ground truth programs are unavailable. The experimental results demonstrate the effectiveness of our proposed approach across different datasets, highlighting its adaptability and robustness in motion Q\&A tasks. Our contributions can be summarized as follows: 
\begin{itemize}
    \item  We propose a novel implicit program-guided reasoning framework that unifies reasoning across multiple operation types without relying on predefined functional modules.

    \item We introduce a program-guided reading mechanism that enables the model to dynamically choose the most suitable features for each step of reasoning.

    \item We achieve state-of-the-art performance on the Babel-QA benchmark and introduce a new motion Q\&A dataset based on HuMMan to evaluate generalization capability.
\end{itemize}

\section{Related Work}
\label{sec:previous_work}

\paragraph{Human Motion Understanding}
\chinthani{Human motion understanding has been widely studied, particularly in action recognition \cite{liu2017pku, asghari2020dynamic, caetano2019skelemotion, cai2021jolo, cheng2020skeleton, choutas2018potion}.   
ST-GCN \cite{yan2018spatial} introduced graph convolutional networks (GCNs) to model spatial-temporal relationships between body joints, establishing a dominant paradigm in skeleton-based action recognition. Subsequent works have improved GCN architectures to enhance accuracy.
Beyond GCNs, PoseConv3D \cite{duan2022revisiting} proposed a 3D heatmap volume representation, enabling 3D-CNN models to achieve superior performance. For untrimmed motion sequences, temporal convolutional networks (TCN) \cite{filtjens2022skeleton, yao2018efficient} and transformer-based models \cite{sun2022locate} estimate per-frame action probabilities to facilitate action localization.
Further, motion-language research has gained attention in linking natural language with 3D human motion. Notable efforts include various motion-language datasets such as KIT \cite{plappert2016kit}, HumanML3D \cite{guo2022generating}, HuMMan \cite{humman}. Existing motion-language models focus mainly on contrastive learning to align motion and text representations \cite{motionpatches, tevet2022motionclip}. 
While these models provide valuable insights into motion sequences, they lack the ability to perform multi-step reasoning that combines action-level understanding with fine-grained motion analysis, such as identifying body parts involved in specific frames or detecting sudden changes in movement. To address this limitation, the recent work NSPose \cite{nspose} introduced the human motion question-answering (QA) task, which evaluates fine-grained human behavior understanding by pairing motion sequences with natural language questions and categorical answers.
Building on this foundation, our study also addresses the human motion QA problem, aiming to enhance human motion reasoning by developing a more adaptive and scalable reasoning framework.}

\paragraph{Reasoning Approaches for Question Answering}
\chinthani{
Question Answering (QA) in Artificial Intelligence is a fundamental research area focused on developing systems capable of understanding and responding to user queries across various modalities, including text, images, video, motion, and multimodal inputs. Early QA systems were mainly built on rule-based or statistical methods \cite{leite2023rules}. However, with the advent of deep learning, neural reasoning has become a dominant approach.
Recent advances in neural reasoning for QA span multiple modalities, including text-based QA \cite{yasunaga2021qa, mirzaee2021spartqa}, image-text QA \cite{malinowski2015ask, ma2016learning, lin2024vila, li2022blip}, and video-text QA \cite{kim2021self, zhang2024video}. Most of these models are built upon Recurrent Neural Networks (RNNs) and Transformer architectures.
Meanwhile, neuro-symbolic reasoning has demonstrated significant success in visual reasoning tasks \cite{yi2018neural, mao2019neuro, NEURIPS2024_f9668d22}. Neuro-symbolic VQA \cite{yi2018neural} integrates symbolic program execution with visual recognition to improve question answering performance, achieving strong performance on the CLEVR benchmark \cite{johnson2017clevr}. NS-CL \cite{hsu2023ns3d} further improved this paradigm by eliminating the need for dense supervision, enabling the model to learn reasoning patterns by observing images and reading natural language questions and answers. MMN \cite{chen2021meta} improves over Neural Module Network \cite{andreas2016neural} by taking in function recipes and morph into diverse instance modules dynamically. 
More recently, neuro-symbolic frameworks have been extended to temporal reasoning \cite{chen2021grounding}, knowledge-based reasoning~\cite{nguyen2025neuro} and 3D reasoning \cite{hong20223d, hsu2023ns3d} as well. 
In human-motion reasoning, NSPose employs a neuro-symbolic reasoning model that explicitly defines functional modules to process motion data. While this approach offers advantages such as interpretability and data efficiency, its reliance on manually defined functional modules poses challenges in terms of scalability and adaptability, particularly when handling diverse and complex motion scenarios. 
Overcoming the above limitations, we propose a neural-based implicit program-guided reasoning approach for human motion Q\&A. Unlike explicit neuro-symbolic methods, our approach learns to reason implicitly without predefined functional operations, allowing for greater flexibility, scalability, and adaptability in complex motion reasoning tasks.}

\section{Method}
\label{sec:method}

\subsection{Overview}

We propose a novel memory mechanism-based reasoning framework for the human motion Q\&A task. This framework is designed to utilize structured program functions to guide the reasoning process, eliminating the need for manually defined functional modules. This approach enables interpretable and flexible reasoning over human motion sequences. The overall framework is illustrated in Fig.~\ref{fig:Architecture} (a). Given a human motion sequence represented by 3D joint locations as $\mathcal{M} \in \mathcal{R}^{T \times J \times 3}$, where $T$ is the number of frames and $J$ denotes the number of joints per frame, and a natural language question $\mathcal{Q}$, the objective is to predict an accurate answer by reasoning over the provided motion sequence. The motion sequence is first processed by a motion encoder, which extracts feature embeddings representing spatial-temporal dynamics. Additionally, we assume the availability of a structured program $\mathcal{P}$, which can either be a ground-truth program or a predicted program from an existing program predictor \cite{dong2018coarse}. Rather than executing directly over the feature representation in a rule-based manner as done in \cite{nspose}, we have developed a unified reasoning module inspired by the MAC architecture \cite{mac}. The original MAC model generates operations by computing soft attention over question words, which introduces potential ambiguity due to language variability and contextual dependencies. To mitigate these limitations, we introduce a program-guided reasoning module. The program guidance ensures that each reasoning step follows an explicitly defined program function, hence eliminating inconsistencies introduced by soft-attention-based reasoning. This reasoning process is further elaborated in Section \ref{sec:program_guided_reasoning_module}. Additionally, different types of questions require reasoning over different motion attributes. For instance, queries about action types demand a high-level semantic understanding of motion sequences, while queries about body parts necessitate fine-grained spatial attention to localized movements. To effectively connect these hierarchical representations, we introduce a program-guided reading mechanism that dynamically selects the most informative motion features for each reasoning step. 


\subsection{Preliminaries on MAC Network}


The Memory-Attention-Control (MAC) network is a neural reasoning architecture designed for compositional reasoning in question-answering tasks. The MAC network operates iteratively, decomposing complex reasoning into multiple stepwise operations, where each step refines an intermediate memory state to arrive at the final answer.
A MAC network consists of three key components:
1) Control Unit which determines the reasoning operation at each step by attending to relevant parts of the question using a soft attention mechanism;
2) Read Unit which extracts pertinent information from the visual input based on the control signal, allowing iterative refinement of retrieved evidence;
3) Write Unit which integrates the extracted information into a recurrent memory state, progressively enhancing the representation of the answer.

The iterative reasoning of the MAC network enables multi-step inference, making it particularly suitable for tasks requiring compositional logic, such as visual question answering (VQA). However, despite its effectiveness in structured reasoning, MAC generates operations based on soft attention over question words, which introduces ambiguity and variability in operation at each reasoning step.

\subsection{Implicit 
Program-Guided Reasoning Module}
\label{sec:program_guided_reasoning_module}
Given the motion sequence $\mathcal{M}$ and question $\mathcal{Q}$, we first use a motion encoder $\mathbf{E}_\text{m}$ and a text encoder $\mathbf{E}_\text{t}$ to extract feature embeddings for the motion sequence and question respectively. We adopted the pre-trained ViT model from \cite{motionpatches} as the motion encoder, which introduces a patch-based representation for the skeleton structure. Instead of considering each joint, the patch-based representation divides the skeleton joints into body parts, hence better encoding skeleton structure information.  The RoBERTa representation in \cite{liu2019roberta} is used as our text encoder. The motion feature $f_\text{m} = \mathbf{E}_\text{m}(\mathcal{M})$ and the text feature $f_\text{t} = \mathbf{E}_\text{t}(\mathcal{Q})$ are first fused to obtain a text-aware motion feature:
\begin{equation}
    h_\text{m} = \text{Attention}(Q=f_\text{m}, K=f_\text{t}, V=f_\text{t}).
\end{equation}
We use cross-attention with the query being the motion feature and the key and value being the text feature.  
In addition to fusing with the text, the motion feature is further fused with the question type information using the same attention mechanism, with query being $h_m$ and key and value being the question type. The question type consists of query direction, query action and query body part, and is encoded with the RoBERTa before fusion.  
The final fused motion feature is used for the iterative reasoning process.

We use the program as the guidance for reasoning. The program is represented as $\mathcal{P} \in \mathcal{R}^{p \times d}$, where $p$ denotes the length of the program and $d$ the program embedding dimension.  Note that the program can be ground truth as used in \cite{nspose} or predicted from an existing program predictor \cite{dong2018coarse}. At each reasoning step, we gradually refine the intermediate memory states guided by the program. 
Specifically, at the $(i+1)$-th reasoning step with previous memory states being $\{\mathcal{S}_1, \mathcal{S}_2...\mathcal{S}_{i}\}$, 
we first attend to the previous states that are related to the current reasoning step by an attention mechanism:
\begin{equation}
    \mathcal{S}'_i = \text{Attention} (Q=P_{i+1}, K=\mathcal{S_{\|B\|}}, V=\mathcal{S_{\|B\|}}),
    \label{eqn:attention_to_pre}
\end{equation}
where $P_{i+1}$ represents the program function for the $(i+1)$-th reasoning step and $\mathcal{S_{\|B\|}}$ represents the previous memory states that are related to the current step. For example, the question ``What action does the person do before they move left?'' and the corresponding program ``\textit{query}\_\textit{action}(\textit{relate}(before, \textit{filter}(left)))'', the program function \textit{relate}() depends on the output of \textit{filter}(). It is important to encode this dependency information to provide context information for each reasoning step. The dependency relationship is provided by the structure information of the program.

In addition to the previous context information, the model also requires information to be extracted from the input motion sequence to conduct reasoning. This process is similar to the reading stage of the MAC, where the key is to extract the most relevant information for the current reasoning step. We achieve this by directly taking the guidance from program functions. For example, for the question `` What action does the person do before they move left?'', the program functions \textit{filter}(move left) or \textit{relate}(before) provides clear information on which part to attend.  This process is implemented as an attention module with $\mathcal{S}'_i$ being the query and motion feature $h_\text{m}$ being the key and query:
\begin{equation}
    \mathcal{S}_{i+1} = \text{Attention} (Q=\mathcal{S}'_i, K=h_\text{m}, V=h_\text{m}).
    \label{eq:program-guided-reading}
\end{equation}
 The intuition is to refine the current memory states $\mathcal{S}'_i$ with relevant motion features in $h_\text{m}$.
 Compared to operation generated by soft attention over question words in the control unit of MAC, the direct program information provides a more precise representation.  As a consequence, the proposed program-guided reasoning mitigates operation ambiguities by directly conditioning each reasoning step on explicit program functions.

\subsection{Program-Guided Reading Mechanism}
\label{sec:program_guided_reading}

We introduce a program-guided reading mechanism to enhance the reasoning capacity of our model by enabling dynamic retrieval from a knowledge pool. Unlike the MAC model, which extracts information from a single static feature representation, our model constructs a multi-level knowledge pool by aggregating intermediate representations from different stages of the Vision Transformer (ViT). This allows the reasoning module to adaptively select the most relevant feature representation at each step based on the program function. We define the knowledge pool as a collection of hierarchical feature embeddings extracted from layers of the ViT model $F_\text{m} = \{h_i\}_{i=1}^{M}$,
where $h_i$ represents the feature embeddings from the $i$-th intermediate layer of ViT. Since feature representations from different layers encode varying levels of abstraction, lower-layer features capture fine-grained local motion details, while higher-layer features encode global semantic information. Different levels of features are important for different motion concepts. For example, the question ``What body part does the person use after they jump''  requires both high-level semantic understanding for the concept `jump' and low-level local information for moving `body part'.  
We apply a projection layer to ensure all feature embeddings reside in a shared latent space:
\begin{equation}
h_i' = W_i^p h_i, \quad \forall i \in {1, ..., M},
\end{equation}
where $W_i^p$ is the projection layer for the $i$-th intermediate output. 
Our program-guided reading mechanism is implemented by replacing the single motion feature $h_\text{m}$ in Eqn.\eqref{eq:program-guided-reading} with the multi-level feature $F'_{\text{m}} = \{h'_i\}_{i=1}^{M}$, expressed as 
\begin{equation}
    \mathcal{S}_{i+1} = \text{Attention} (Q=\mathcal{S}'_i, K=\{h_i'\}_{i=1}^{M}, V=\{h_i'\}_{i=1}^{M}).
\end{equation}
We also illustrate the detailed process of one reasoning step in Fig.~\ref{fig:Architecture} (b). By incorporating the most informative feature representation at each step, our model can effectively capture dependencies across hierarchical motion features, making it highly adaptable to diverse motion concepts.

\subsection{Iterative Reasoning}
\label{sec:interative_reasoning}

The reasoning module executes for $p$ iterative steps, progressively refining the memory state until reaching the final representation. Each intermediate state is stored in a memory matrix, allowing the model to leverage dependency information between reasoning steps, as formulated in Eqn~\eqref{eqn:attention_to_pre}. The iterative refinement process is formulated as:

\begin{equation}
\mathcal{S}_i = f(\mathcal{S}_{i-1}, \mathcal{P}_i, F'_\text{m}), \quad i = 1, 2, ..., p,
\end{equation}
where $f$ represents the reasoning function integrating prior memory $\mathcal{S}_{i-1}$, the program function $\mathcal{P}_i$, and motion features $F_\text{m}$.
Once the final reasoning step is completed, the resulting representation is passed into a \textbf{multi-branch classifier}, where each branch specializes in predicting answers corresponding to a specific question type, including query\_action, query\_direction, and query\_body\_part.
For a given question, only the corresponding branch depending on the question type is activated.
The question type information can be directly obtained from the provided program. 

\subsection{Training and Optimization}
\label{sec:train_and_optimization}
The objective of our model is to predict the correct answer based on the input motion sequence and question. We employ a classification-based answer prediction framework, where the model predicts a probability distribution over the candidate answers in the dataset. The predicted answer vector $v^{\text{ans}} \in \mathcal{R}^{a}$ represents the confidence scores for $a$ possible answers.
The loss function of our model can be defined as:
\begin{equation}
    \mathcal{L} = -\log p(v^{\text{ans}} | \mathcal{M}, \mathcal{Q}, \mathcal{P}).
    \label{eqn:loss function}
\end{equation}
We use the AdamW optimizer with weight-decay regularization to prevent over-fitting, the weight-decay coefficient is set to 0.0001.

\section{Experiments}
\label{sec:experiments}

\paragraph{Training Details}
\chinthani{
We used the pre-trained motion ViT encoder in \cite{motionpatches} to extract the multilevel motion features required for the program-guided reading mechanism of our IMoRe. 
This converts raw motion sequences into a unified representation known as motion patches, which serve as the input for the ViT-based motion encoder, with each patch encapsulating joints relevant to specific body parts such as the torso, right leg, and left arm. 
We used the RoBERTa \cite{liu2019roberta} text encoder to extract the required text embeddings.
The network was trained for 100 epochs using the AdamW optimizer with a learning rate of 1e-6, dropout of 0.1, and batch size 4 on a 48 GB NVIDIA RTX 6000 GPU. All experiments were repeated three times, and the average results were reported.}

\paragraph{Implementation Details}
We implement our approach under two settings as, 
\textbf{(1) IMoRe I:} In this setting, we provide the entire human motion sequence as input. Since the motion ViT encoder supports segments of 224 frames, we partition the sequence accordingly and pass each segment separately through the encoder covering the entire motion sequence. 
\textbf{(2) IMoRe II:} In this setting, rather than processing the complete sequence, we randomly sample 224 frames from the whole sequence with a random starting index following \cite{motionpatches}. During validation and testing, we execute five runs for the same sequence and select the run with the highest logits scores. The intuition behind this setting is that the segment with the highest score should include the segment with the query concept. 
We construct the composite knowledge pool required for the program-guided reading mechanism described in Section \ref{sec:program_guided_reading} by extracting features from $M=6$ intermediate layers (i.e. $i \in \{0,2,4,6,8,11\}$) of motion ViT along with its final encoder output. Consequently, each 224-frame segment is processed through these six intermediate layers and the final encoder to form the composite knowledge pool.

We further evaluate our approach under two scenarios: one with ground truth programs provided and one with predicted programs. For the latter scenario, we adapt the coarse-to-fine two-stage program generation network from \cite{dong2018coarse} to predict the required programs. This experiment is to show the robustness of our method to noisy programs since the ground truth program is not always available in practice. Experimental results denoted with an asterisk mark (*) representing the performance with predicted programs.

\paragraph{Datasets.}
\chinthani{
We evaluate our model on two datasets Babel-QA \cite{nspose} and HuMMan-QA. 
\textbf{(1) Babel-QA}: consists of 1800 training questions, 384 validation questions, and 393 test questions categorized into action, direction, and body part-related questions extracted from 771 motion sequences.
\textbf{(2) HuMMan-QA}: Given the small size of the Babel-QA dataset and our goal to test the generalizability of our model across diverse datasets, we developed the HuMMan-QA dataset based on the HuMMan-MoGen dataset \cite{humman}. The HuMMan-MoGen dataset comprises 6,264 SMPL motion sequences from 160 actions, each annotated with 112,112 detailed temporal and spatial text descriptions.
We utilized the motion QA generation strategy in \cite{nspose}.
However, since HuMMan-MoGen lacks segment-level labels for actions, directions, and body parts, we employed the GPT-4o model with in-context learning examples to generate these labels, which were then meticulously refined and manually verified against the actual motion sequences by the authors. 
The HuMMan-QA dataset consists of 2066 training questions, 524 validation questions, and 533 test questions, with 158, 18, and 6 classes for action, body parts, and direction, respectively, derived from 1,311 motion sequences.
Accuracy is used as the evaluation metric.}

\subsection{Results on Babel-QA Dataset}

We report the results of our IMoRe on the Babel-QA dataset in Table \ref{tab:BABEL-QA results}. We compare our approach with state-of-the-art approaches, including NSPose \cite{nspose}, which is a neural symbolic approach with pre-define functional modules, and CLIP, 2S-AGCN-MLP, 2S-AGCN-RNN, MotionCLIP-MLP, and 
MotionCLIP-RNN \cite{nspose}, which directly fuses motion and text features. 
The results reveal several key insights. Firstly, the neural symbolic-based approaches (NSPose, IMoRe) have better performance compared with direct feature fusion based approaches (2S-AGCN-MLP, 2S-AGCN-RNN, MotionCLIP-MLP,
MotionCLIP-RNN). This shows the effectiveness of the neural symbolic based approach for the motion reasoning task. 
Secondly, it can be seen that IMoRe I and IMoRe II outperform the state-of-the-art NSPose \cite{nspose} by a significant margin for the ground truth and predicted program setting. This demonstrates the effectiveness of our implicit program-guided reasoning approach for human motion reasoning. 
Lastly, while our IMoRe experiences only a modest performance drop (1-2\%) in the predicted program setting, NSPose exhibits a substantial decline in performance. We attribute this to the fundamental differences between explicit and implicit reasoning approaches. In explicit reasoning, functional programs are highly dependent on the correct execution of each step: an error in any predicted step can cascade into a significant performance drop. In contrast, implicit reasoning provides greater adaptability, allowing the model to recover from incorrect input signals and mitigate performance degradation. This resilience highlights the robustness of our implicit approach in handling noisy or imperfect program predictions, making it more suitable for real-world scenarios where execution errors are inevitable.

\begin{table*}[ht]
  \centering
  \resizebox{0.95\textwidth}{!}{
  \begin{tabular}{|l|c|c|c|c|c|c|c|c|c|c|c|c|c|}
    \toprule
    \multirow{2}{*}{Model} 
    & \multirow{2}{*}{Overall} 

    & \multicolumn{4}{c|}{Query action} 
    & \multicolumn{4}{c|}{Query direction} 
    & \multicolumn{4}{c|}{Query body part} \\

    \cline{3-14}
    
    & & \multicolumn{1}{c|}{All} & \multicolumn{1}{c|}{Before} &  \multicolumn{1}{c|}{After} & \multicolumn{1}{c|}{BTW}
    & \multicolumn{1}{c|}{All} & \multicolumn{1}{c|}{Before} &  \multicolumn{1}{c|}{After} & \multicolumn{1}{c|}{BTW}
    & \multicolumn{1}{c|}{All} & \multicolumn{1}{c|}{Before} &  \multicolumn{1}{c|}{After} & \multicolumn{1}{c|}{BTW} \\

    \toprule

    CLIP \cite{Radford2021LearningTV}         & 0.417 & 0.467 & 0.380 & 0.452 & 0.591 & 0.366 & 0.467 & 0.292 & 0.222 & 0.261 & 0.261 & 0.278 & \underline{0.333} \\
    
    2S-AGCN-MLP \cite{nspose}     & 0.355 & 0.384 & 0.353 & 0.411 & 0.273 & 0.352 & 0.378 & 0.250 & 0.278 & 0.228 & 0.261 & 0.130 & \underline{0.333} \\
    
    2S-AGCN-RNN \cite{nspose}     & 0.357 & 0.396 & 0.349 & 0.396 & 0.409 & 0.352 & 0.400 & 0.396 & 0.278 & 0.194 & 0.261 & 0.111 & 0.167 \\
    
    MOTIONCLIP-MLP \cite{nspose}  & 0.430 & 0.485 & 0.411 & 0.470 & 0.545 & 0.361 & 0.400 & 0.271 & 0.333 & 0.272 & 0.304 & 0.222 & \underline{0.333} \\
    
    MOTIONCLIP-RNN \cite{nspose}  & 0.420 & 0.489 & 0.461 & 0.441 & 0.606 & 0.310 & 0.400 & 0.333 & 0.222 & 0.250 & 0.333 & 0.167 & \underline{0.333} \\
    \hline

    NSPOSE \cite{nspose}         & 0.578 & 0.627 & 0.618 & 0.620 & 0.639 & 0.598 & 0.389 & \textbf{0.583} & \underline{0.750} & 0.325 & 0.296 & \textbf{0.471} & 0.083 \\

    NSPOSE* \cite{nspose}        & 0.472 & 0.519 & 0.570 & 0.493 & 0.500 & 0.488 & 0.375 & 0.238 & 0.375 & 0.189 & 0.250 & 0.176 & 0.167 \\
    
    \hline

    IMoRe I & 0.609 & 0.652 & \underline{0.640} & 0.676 & \underline{0.722} & 0.622 & 0.486 & 0.393 & 0.458 & \underline{0.373} & \underline{0.389} & 0.353 & \textbf{0.440} \\
    
    IMoRe I* & 0.602 & 0.646 & \underline{0.640} & 0.671 & 0.713 & 0.611 & \underline{0.514} & 0.357 & 0.375 & 0.351 & 0.380 & 0.343 & 0.250 \\
    
    IMoRe II & \textbf{0.640} & \textbf{0.695} & \textbf{0.677} & \textbf{0.707} & \textbf{0.750} & \textbf{0.679} & 0.458 & \underline{0.560} & \textbf{0.792} & 0.358 & \textbf{0.407} & \underline{0.441} & 0.083 \\
    
    IMoRe II* & \underline{0.615} & \underline{0.656} & 0.610 & \underline{0.700} & 0.667 & \underline{0.624} & \textbf{0.528} & 0.429 & 0.583 & \textbf{0.391} & \underline{0.389} & \underline{0.441} & \underline{0.333} \\

   \bottomrule
   
  \end{tabular}%
  }
  \caption{Evaluation of IMoRe and existing models on Babel-QA test set. The asterisk mark (*) refers to the setting where the model uses the predicted programs as the input. BTW refers to `Between'. The bold and underline font shows the best and the second best result respectively.}
  \label{tab:BABEL-QA results}
\end{table*}

\subsection{Generalization to HuMMan-QA Dataset}

We further evaluate the generalizability of our approach on the HuMMan-QA dataset, with results presented in Table \ref{tab:HuMMan-QA results}. We can see that both IMoRe I and IMoRe II consistently outperform existing models, including both neural symbolic-based approach NSPose and direct motion-text feature fusion-based approach MotionCLIP-MLP and MotionCLIP-RNN, achieving significant performance improvements of 4\% and 5.6\%, respectively. This demonstrates the ability of our network to generalize effectively across different datasets.
Additionally, we observe that while MotionCLIP-MLP does not perform as well on the Babel-QA dataset, it achieves competitive results on HuMMan-QA. This suggests that MotionCLIP-MLP might be particularly suited for the characteristics of the HuMMan-QA dataset, highlighting the importance of dataset-specific model adaptability. These results highlight the effectiveness of our program-guided implicit reasoning approach and its capacity to generalize to different motion question-answering benchmarks. It should be noted that performance across all models in `Direction: BTW' type questions is zero because no question is generated for this type for the HuMMan-QA Dataset. 
\begin{table*}[t]
  \centering
  \resizebox{0.95\textwidth}{!}{
  \begin{tabular}{|l|c|c|c|c|c|c|c|c|c|c|c|c|c|}
    \toprule
    \multirow{2}{*}{Model} 
    & \multirow{2}{*}{Overall} 

    & \multicolumn{4}{c|}{Query action} 
    & \multicolumn{4}{c|}{Query direction} 
    & \multicolumn{4}{c|}{Query body part} \\

    \cline{3-14}
    
    & & \multicolumn{1}{c|}{All} & \multicolumn{1}{c|}{Before} &  \multicolumn{1}{c|}{After} & \multicolumn{1}{c|}{BTW}
    & \multicolumn{1}{c|}{All} & \multicolumn{1}{c|}{Before} &  \multicolumn{1}{c|}{After} & \multicolumn{1}{c|}{BTW}
    & \multicolumn{1}{c|}{All} & \multicolumn{1}{c|}{Before} &  \multicolumn{1}{c|}{After} & \multicolumn{1}{c|}{BTW} \\

    \toprule

    MotionCLIP-MLP \cite{nspose}  & 0.686 & 0.682 & 0.640 & 0.628 & 0.467 & 0.750 & \underline{0.500} & \textbf{1.000} & 0.000 & \underline{0.692} & \textbf{0.696} & 0.644 & 0.639 \\
    
    MotionCLIP-RNN \cite{nspose}  & 0.623 & 0.601 & 0.559 & 0.541 & 0.583 &
    0.750 & \textbf{1.000} & 0.500 & 0.000 &
    0.667 & \underline{0.638} & 0.616 & 0.694\\
    
    NSPose \cite{nspose} & 0.691 & 0.700 & \textbf{0.686} & 0.610 & 0.729 & \underline{0.822} & 0.425 & \underline{0.833} & 0.000 & 0.677 & 0.620 & 0.639 & 0.833 \\
    
    IMoRe I & \underline{0.719} & \underline{0.744} & \underline{0.652} & \underline{0.734} & \textbf{0.854} & \textbf{1.000} & \textbf{1.000} & \textbf{1.000} & 0.000 & 0.665 & 0.609 & \underline{0.647} & \textbf{0.889} \\
    
    IMoRe II & \textbf{0.730} & \textbf{0.746} & 0.648 & \textbf{0.739} & \underline{0.813} & \textbf{1.000} & \textbf{1.000} & \textbf{1.000} & 0.000 & \textbf{0.717} & 0.636 & \textbf{0.703} & \underline{0.861} \\

   \bottomrule
   
  \end{tabular}%
  }
  \caption{Evaluating the generalizability of our approach on HuMMan-QA dataset. BTW refers to `Between'. There is no question generated for `Between' relation for query direction in this dataset. The bold and underline font shows the best and the second best result respectively. }
  \label{tab:HuMMan-QA results}
\end{table*}


\subsection{Ablation Study}

We perform an ablation study on the Babel-QA dataset to verify the effectiveness of each component. We evaluate four models:
A. ViT + explicit reasoning (NSPose); B. ViT + implicit reasoning (MAC);
C. Implicit program-guided reasoning;
D. Implicit Program-guided reasoning + feature selection (i.e., our full model). The first model replaces the motion encoder used in NSPose with ViT to show that our performance gain is not coming from a different motion encoder.  The second model is to show the effectiveness of implicit reasoning. 
We re-implemented the MAC \cite{mac} network on the Babel-QA dataset.
The third model demonstrates the benefit of program guidance, while the fourth highlights the advantage of program-guided feature selection.

The results are shown in Table \ref{tab:BABEL-QA_ablation}, from which we can draw several conclusions. First, the performance of NSPose drops when replacing the encoder with the ViT encoder, which shows that the better performance of our IMoRe comes from a better reasoning module. Second, implicit reasoning yields improved results compared to explicit reasoning. Despite the simplicity of MAC’s design, its implicit reasoning mechanism surpasses NSPose's explicit reasoning. This advantage likely stems from the inherent adaptability of implicit reasoning, which allows the model to handle diverse reasoning tasks more effectively without relying explicitly on predefined function modules. Third, the better performance of program-guided reasoning compared to MAC verifies the effectiveness of the program guidance. 
Third, the incorporation of program-guided feature selection leads to notable performance improvements of 5.5\%. This highlights the advantage of extracting task-relevant motion features based on the program information. For instance, action-type queries benefit from high-level motion features, while body-part-related queries require fine-grained motion representations. By dynamically selecting relevant features, our implicit reasoning with program-guided feature selection effectively tailors feature extraction to the specific reasoning task.

\begin{table*}[h!]
  \centering
  \resizebox{0.9\textwidth}{!}{
  \begin{tabular}{|l|c|c|c|c|c|c|c|c|c|c|c|c|c|}
    \toprule
    \multirow{2}{*}{Model} 
    & \multirow{2}{*}{Overall} 

    & \multicolumn{4}{c|}{Query action} 
    & \multicolumn{4}{c|}{Query direction} 
    & \multicolumn{4}{c|}{Query body part} \\

    \cline{3-14}
    
    & & \multicolumn{1}{c|}{All} & \multicolumn{1}{c|}{Before} &  \multicolumn{1}{c|}{After} & \multicolumn{1}{c|}{BTW}
    & \multicolumn{1}{c|}{All} & \multicolumn{1}{c|}{Before} &  \multicolumn{1}{c|}{After} & \multicolumn{1}{c|}{BTW}
    & \multicolumn{1}{c|}{All} & \multicolumn{1}{c|}{Before} &  \multicolumn{1}{c|}{After} & \multicolumn{1}{c|}{BTW} \\

    \toprule

    A. ViT + NSPose \cite{nspose} & 0.440 & 0.438 & 0.433 & 0.452 & 0.211 & 0.450 & 0.300 & 0.270 & 0.000 & \textbf{0.362} & \textbf{0.476} & 0.258 & \textbf{0.667} \\
    
    B. ViT + MAC \cite{mac} & 0.582 & 0.649 & 0.597 & 0.633 & \underline{0.712} &
    0.526 & 0.356 & \underline{0.458} & \underline{0.500} & 
    0.356 & 0.406 & \underline{0.315} & \underline{0.083} \\
    
    
    
    
    C. Program-guided reasoning & \underline{0.607} & \underline{0.676} & \underline{0.632} & \underline{0.697} & 0.704 & \underline{0.571} & \underline{0.403} & 0.321 & \underline{0.500} & 0.340 & 0.352 & \textbf{0.441} & \underline{0.083} \\
    
    D. Program-guided reasoning + feature selection & \textbf{0.640} & \textbf{0.695} & \textbf{0.677} & \textbf{0.707} & \textbf{0.750} & \textbf{0.679} & \textbf{0.458} & \textbf{0.560} & \textbf{0.792} & \underline{0.358} & \underline{0.407} & \textbf{0.441} & 0.000 \\


   \bottomrule
   
  \end{tabular}%
  }
  \caption{Ablation study on Babel-QA dataset for explicit reasoning, implicit reasoning, implicit reasoning without program-guided feature selection, and implicit reasoning with program-guided feature selection. BTW refers to `Between'. The bold and underline font shows the best and the second best result, respectively.}
  \label{tab:BABEL-QA_ablation}
\end{table*}

\begin{figure*}[h!]
\centering
\includegraphics[width=1\textwidth]{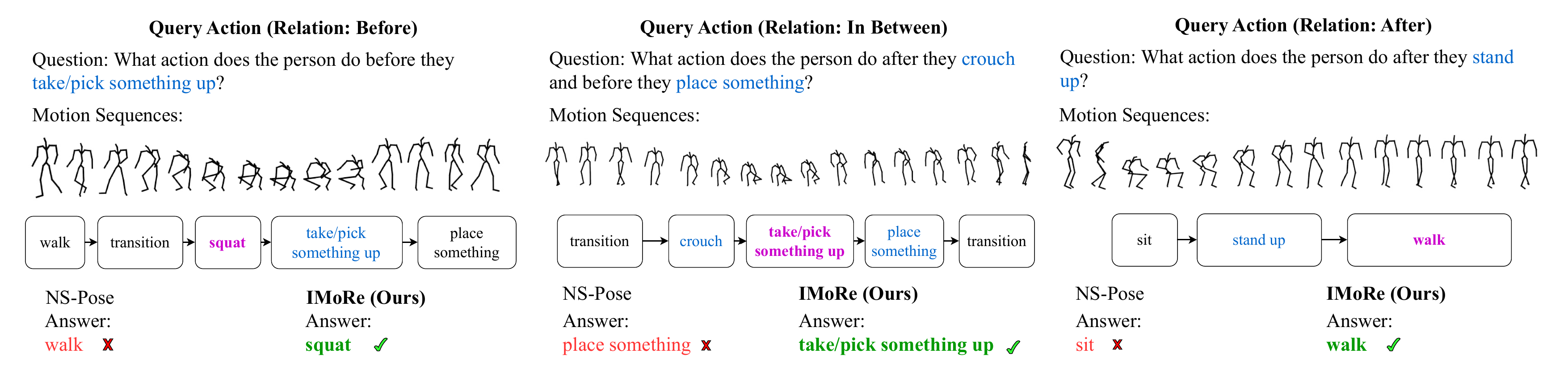}
\vspace{2mm}
\includegraphics[width=1\textwidth]{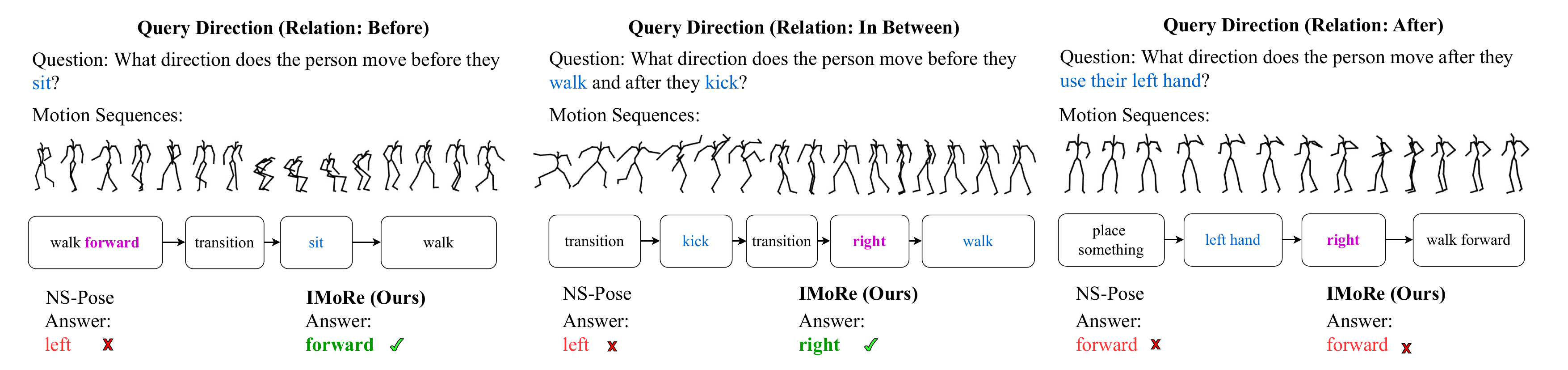}
\vspace{2mm}
\includegraphics[width=1\textwidth]{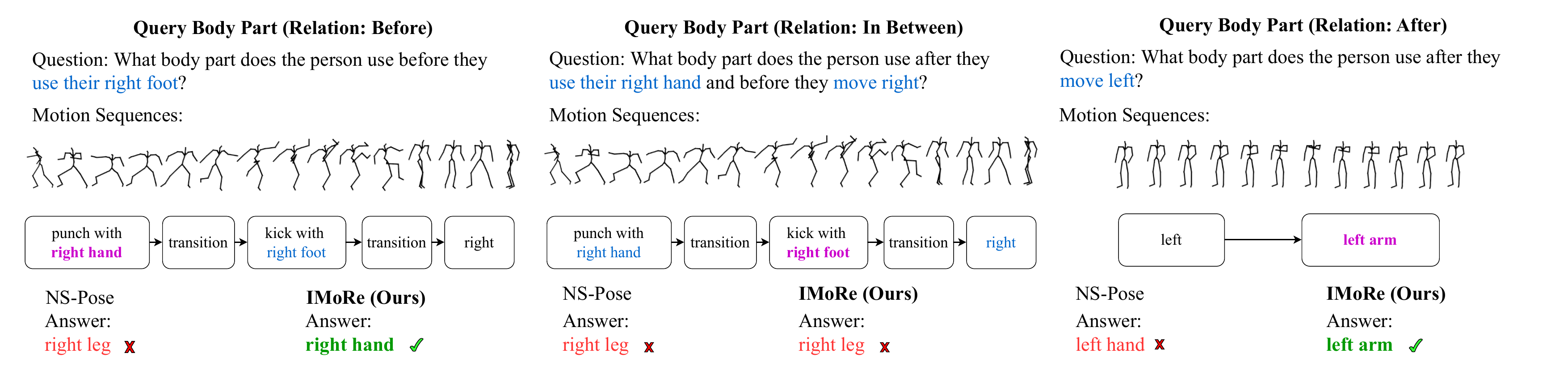}


\caption{
    Qualitative comparison between our method (IMoRe) and NSPose on motion sequence reasoning tasks with different query types: action, direction, and body part with different temporal relations including before, in between, and after.
    }
\label{fig:qualitative}
\end{figure*}

\subsection{Qualitative Results}
The qualitative examples in Fig.~\ref{fig:qualitative} illustrate the effectiveness of our proposed IMoRe method in understanding motion sequences with relational reasoning across different aspects: action, direction, and body part. 
Compared to NSPose, our model consistently provides more accurate reasoning results across various relational queries.  
In the first example (top left), when asked what action occurs before ``taking/picking something up'', IMoRe correctly predicts ``squat,'' whereas NSPose incorrectly outputs ``walk'' which occurred earlier in the sequence.
This highlights IMoRe’s ability to capture fine-grained, preparatory actions that are crucial for understanding human motion.
In the example of query direction in between two movements, while NS-Pose wrongly predicts ``left'', IMoRe correctly predicts ``right'' showcasing its ability to model spatial transitions.
In the query body part ``before'' relation case, IMoRe correctly identifies that the ``right hand'' is used before the ``right foot'', while NS-Pose incorrectly predicts the ``right leg''. 
This indicates that IMoRe effectively captures body part dependencies by dynamically attending to relevant motion frames, whereas NS-Pose fails to focus on the correct segment.
Despite its strong performance, IMoRe still encounters challenges in certain cases.  
One notable failure occurs in queries involving body parts with the ``in-between'' relation, where the model struggles to differentiate between closely related actions.
For example, when asked about the body part used after one action and before another, IMoRe struggles likely due to the confusion of similar concepts.
Another failure case appears in movement direction queries with the ``after'' relation, where both IMoRe and NSPose incorrectly infer the \textit{forward} direction instead of recognizing the subtle directional change present in the motion sequence.

\section{Conclusion}
\label{sec:conclusion}
We presented an implicit program-guided reasoning framework for human motion Q\&A, addressing the limitations of explicit program execution. Our method dynamically adapts reasoning based on structured program functions to improve scalability and adaptability. The program-guided reading mechanism enables the dynamic selection of multi-level motion features from a pre-trained ViT model, allowing the model to reason over both high-level semantics and fine-grained motion cues.
Our model achieves state-of-the-art performance on the Babel-QA dataset and generalizes to the newly constructed HuMMan-QA dataset. 
Analysis with predicted programs confirms the robustness and effectiveness of our approach in addressing diverse motion-related queries in real-world scenarios.
\vspace{1mm}

\noindent
\textbf{Acknowledgments}
This research/project is supported by the National Research Foundation, Singapore, under its NRF Fellowship (Award\# NRF-NRFF14-2022-0001) and by funding allocation to Basura Fernando by the A*STAR under its SERC Central Research Fund (CRF), as well as its Centre for Frontier AI Research.
\newpage
{
    \small
    \bibliographystyle{ieeenat_fullname}
    \bibliography{main}
}

\pagebreak


	\twocolumn[{%
		\renewcommand\twocolumn[1][]{#1}%
		\vskip .5in
		\begin{center}
			\textbf{\Large Supplementary Material for}\\
			\vspace*{8pt}
			\textbf{\Large IMoRe: Implicit Program-Guided Reasoning for Human Motion Q\&A  } \\			
		\end{center}
        \vspace{8mm}
	}]

\setcounter{equation}{0}
\setcounter{figure}{0}
\setcounter{table}{0}
\setcounter{section}{0}
\setcounter{page}{11}

\noindent \textbf{Comparison with video-language models trained on large-scale data.} We finetuned Qwen-2.5-VL-3B, InternVL2.5-4B-MPO and MiniCPM-V2.6 on the Babel-QA dataset where the motion sequences are converted to videos consisting of skeleton images. Program and answer set information are provided in the prompt for fair comparison. As shown in Table \ref{comparison_VLMs}, IMoRe significantly outperforms all VLM baselines. We attribute this to: 
(1) VLMs struggle to capture fine-grained temporal and spatial concepts from motion image sequences, and
(2) their limited ability to leverage structured programs during reasoning.

\begin{table}[h!]
\centering
\small
\resizebox{\columnwidth}{!}{
\begin{tabular}{|c|c|c|c|c|}
\hline
 & Overall & Action & Direction & Bodypart \\
\hline
IMoRe I & 0.609 & 0.652 & 0.622 & 0.373 \\
Qwen-2.5-VL-3B & 0.425 & 0.467 & 0.333 & 0.350 \\
InternVL2\_5-4B & 0.402 & 0.433 & 0.306 & 0.383 \\
MiniCPM & 0.384 & 0.410 & 0.306 & 0.367 \\
\hline
\end{tabular}
}
\caption{Comparison with VLMs}
\label{comparison_VLMs}
\end{table}

\noindent \textbf{Additional ablation study.} (a) Comparison between question and program for text-aware feature. We fuse the motion feature and the text feature to obtain the text-aware feature as described in Sec. 3.3 of the main paper. To verify this design, we show results using programs for text-aware motion feature (IMoRe w Pro) in Table \ref{addition_ablation}. We can see that the performance drops slightly, likely because the question texts better guide token-level attention in ViT-derived motion features, improving the alignment between text and motion. (b) Question type information. We also ablate over the question type information as described in Sec 3.3 of the main paper. We can see that the performance without question type (IMoRe wo QT) in Table \ref{addition_ablation} slightly drops.

\begin{table}[h!]
\centering
\small
\begin{tabular}{|c|c|c|c|c|}
\hline
 & Overall & Action & Direction & Bodypart \\
\hline
IMoRe I & 0.609 & 0.652 & 0.622 & 0.373 \\
IMoRe w Pro & 0.598 & 0.690 & 0.583 & 0.217 \\
IMoRe wo QT & 0.603 & 0.655 & 0.596 & 0.254 \\
\hline
\end{tabular}
\caption{Additional ablation study}
\label{addition_ablation}
\end{table}

\begin{figure}[h!]
  \centering
  \includegraphics[width=0.99\linewidth]{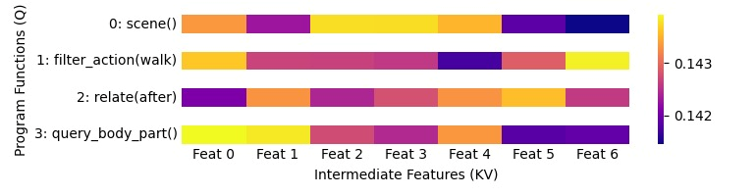}
    \caption{Visualization of feature level selection.}
       \label{fig:feature_selection}
\end{figure}

\noindent \textbf{Visualization of feature level selection.} We have compared single-level and multi-level feature selection quantitatively in the ablation study (C vs D setting in Table 3 of main paper). The improvement from C to D verifies the effectiveness of the multi-level based feature selection. We further show a qualitative attention map between program functions (y-axis) and multi-level features (x-axis) in Fig.~\ref{fig:feature_selection}. It can be seen that \textbf{filter\_action()} attends to high-level feature (Feat 6) while \textbf{query\_body\_part()} to low-level features (Feat 0 and 1). 

\begin{figure}[h!]
  \centering
  \includegraphics[width=0.99\linewidth]{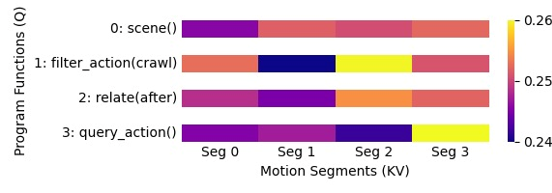}
    \caption{Visualization of concept localization.}
       \label{fig:concept_localization}
\end{figure}

\noindent \textbf{Visualization of concept localization.} Our approach retains interpretability in two ways despite the implicit program guidance. (1) Program structure: structured programs define an explicit step-by-step reasoning path, where each function (\eg, filter, relate) defines a specific operation. Although the reasoning is implemented implicitly, our model follows this program-defined sequence, making each reasoning step directly attributable to a specific symbolic instruction.
(2) Intermediate traceability: As shown in Fig.~\ref{fig:concept_localization}, the attention map between program functions (y-axis) and motion segments (x-axis) reveals interpretable reasoning. After initialization at step 0,  \textbf{filter\_action(crawl)} at step 1 correctly attends to segment 2 (where the action occurs), thereafter attention for \textbf{relate(after)} and \textbf{query\_action()} shifts towards segment 3, illustrating stepwise execution consistent with the program.

\end{document}